# Machine Generalization and Human Categorization: An Information-Theoretic View

James E. Corter  
*Columbia University*

Mark A. Gluck  
*Stanford University*

## 1. INTRODUCTION

In designing an intelligent system that must be able to explain its reasoning to a human user, or to provide generalizations that the human user finds reasonable, it may be useful to take into consideration psychological data on what types of concepts and categories people naturally use. The psychological literature on concept learning and categorization provides strong evidence that certain categories are more easily learned, recalled, and recognized than others. We show here how a measure of the informational value of a category predicts the results of several important categorization experiments better than standard alternative explanations. This suggests that information-based approaches to machine generalization may prove particularly useful and natural for human users of the systems.

## 2. WHAT CONSTITUTES AN OPTIMAL CATEGORY?

### 2.1 Psychological Evidence Concerning the Optimality of Categories

Many studies have shown that some categories or groupings of instances are easier than others to learn and recall as coherent concepts or generalizations. For example, within a hierarchically nested set of categories (such as a taxonomy of animals), there is some level of abstraction--called the "basic level"--that is most natural for people to use (Rosch, Mervis, Gray, Johnson, & Boyes-Braem, 1976). For example, in the hierarchy *animal-bird-robin*, *bird* is the basic level category. The preferential status of basic level categories can be measured in a variety of ways. Basic level names are generally learned earlier by children (Rosch et al., 1976; Daehler, Lonardo, and Bukatko, 1979), and arise earlier in the development of languages (Berlin, Breedlove, & Raven, 1973). People tend to spontaneously name pictured objects at the basic level, and can name them faster at this level than at subordinate or superordinate levels (Rosch et al., 1976; Jolicoeur, Gluck, & Kosslyn, 1984).

---

For their guidance and comments, we are indebted to Gordon Bower, Paul Rosenbloom, Misha Pavel, W. K. Estes, Peter Cheeseman, Ed Smith, Doug Medin, Joachim Hoffmann, and Greg Murphy. The assistance of Katie Albiston and Audrey Weinland is also gratefuly acknowledged. Please address correspondence to: Dr. James E. Corter, Statistics & Measurement, Teachers College-Columbia Universtiy, Box 41, New York, NY, 10027.

## 2.2 Structural Explanations of the Optimality of Categories

Recent findings suggest that the superiority of basic level categories is due to *structural* properties of the categories, that is, to the distribution of features across instances and non-instances (Murphy & Smith, 1982; Hoffmann & Ziessler, 1983). Rosch & Mervis (1975) suggested that basic-level categories are those for which the average *cue validity* of the features for the category is maximal. Cue validity is the extent to which the presence of a feature $f$ predicts the presence of a category $c$, and is generally measured by $p(c|f)$, the conditional probability of the category given the feature. Another possibility is that basic level categories are those for which *category validity* is maximal. Category validity is the converse of cue validity: it represents the extent to which knowing that something is a member of a category enables prediction that it has the feature, and is measured by $p(f|c)$.

There are, however, logical problems with these measures as pointed out by Murphy (1982) and Medin (1983). In a strict hierarchy of concepts, cue validity will always select the most general or inclusive level as optimal, while category validity will tend to select the most specific categories as best. All the types of evidence we mentioned above, such as reaction time to name objects, indicate that it is generally some intermediate level of generalization that is optimal.

A third possibilty, suggested by Jones (1983), is that basic level categories are those that maximize some function *combining* cue and category validity. Jones suggested the product of the two measures as a possible function, and termed this the feature *collocation* measure. In a later section we will examine the performance of all three of these measures in predicting data from certain categorization experiments.

A serious problem with all of these measures is that they are purely extensional, measuring regularities and invariances in the world irrespective of the contexts and needs of the people who are creating and using concepts and categories. We present here an alternative *context-sensitive* measure of the utility of categorizations.

## 3. CATEGORY UTILITY

### 3.1 The Informational Value of Categories

We suggest that the degree to which certain concepts are favored over others may be related to how useful these concepts are for encoding and communicating information about the properties of things in the world. In other words, the most useful categories are those that are, on the average, optimal for communicating information (hence reducing uncertainty) about the properties of instances. We will show how to formalize this idea in situations where the relevant attributes are well defined.

We consider two specific definitions of uncertainty and show the implications of each for Category Utility. First, we utilize the standard definition of uncertainty from information theory (Shannon & Weaver, 1949), and show what it implies about Category Utility. Second, we consider a hypothetical communication game in which one person attempts to transmit information about an item's attributes to another person. Within this game, we interpret uncertainty as an inability to predict attributes, and analyze how category membership information can be used to transmit information about the attributes of objects or events.

We will describe our theory of Category Utility within the context of a finite population of items, each of which is describable in terms of a set of multi-valued nominal attribute dimensions. Each attribute dimension (e.g. *eye color*) is assumed to have a set of possible values (e.g. *green, brown, blue*), one of which occurs in every instance. A category of instances can be described by specifying the *distributions* of attribute values for instances in the category. For example, a specific category of faces may have 40% green eyes, 50% brown eyes, and 10% blue eyes.

In information theory (Shannon & Weaver, 1949), the *uncertainty* of a set, $F$, of $n$ messages (i.e. $F = f_1, f_2, \cdots, f_n$) is given by

$$U(F) = -\sum_{i=1}^{n} P(f_i) \log P(f_i).$$

We consider an attribute dimension to be a set of messages regarding the possible values of the attribute dimension. Consider also a partition, $C$, of a population of objects into two sets: those which are members of a category $c$ and those which are not. Given information that an item is a member of category $c$, the uncertainty of the values of attribute dimension $F$ will be:

$$U(F|c) = -\sum_{i=1}^{n} P(f_i|c) \log P(f_i|c),$$

where $P(f_i|c)$ is the conditional probability that a member of category $c$ has value $f_i$ on attribute dimension $F$. If instances of $c$ occcur with probability $p(c)$ and instances of *not–c* occur with probability $(1-p(c))$, then the expected reduction in uncertainty when one is told the category or not-category information is:

**Category Utility(C,F)**

$$= \left[ P(c) \sum_{i=1}^{n} P(f_i|c) \log P(f_i|c) + (1-P(c)) \sum_{i=1}^{n} P(f_i| \text{ not } c) \log P(f_i| \text{ not } c) \right] - \sum_{i=1}^{n} P(f_i) \log P(f_i).$$

This measure of Category Utility is identical to the standard notion of the *information transmitted* between the message sets C and F.

In certain applications, we may be interested in defining the informational value of category $c$ separately from that of *not–c*. The Category Utility of category $c$ alone is given by:

$$\textbf{Category Utility}(c,F) = P(c) \left[ -\sum P(f_i) \log P(f_i) - -\sum_{i=1}^{n} P(f_i|c) \log P(f_i|c) \right].$$

*3.2 The Guessing-Game Measure of Category Utility*

The information-theoretic measures of Category Utility given in the preceding section have close connections to expected performance in a feature prediction task. If we consider the expected score of someone guessing the values of each attribute dimension of an item, we can compare their expected score when they know nothing about the item to their expected score when they are told whether the items belong to $c$ or $not-c$.

Assuming that the receiver adopts a *probability-matching strategy* (e.g. the receiver guesses value $f_i$ with a probability equal to his expectation of the likelihood of $f_i$ occuring given $c$ or $not-c$) their expected increase in score given the category message can be shown to be given by:

$$\text{Category Utility}(C,F) = \left[ P(c)\sum_{i=1}^{n} P(f_i|c)^2 + (1-P(c))\sum_{i=1}^{n} P(f_i|\ not\ c)^2 \right] - \sum_{i=1}^{n} P(f_i)^2.$$

If the receiver is assumed to have no information about the $not-c$ distribution, then the expected increase in score is

$$\text{Category Utility}(c,F) = P(c)\left[ \sum_{i=1}^{n} P(f_i|c)^2 - \sum_{i=1}^{n} P(f_i)^2 \right].$$

The information-theoretic and expected-score measures of Category Utility are closely related both in mathematical form and in terms of how they order categories as to relative goodness because $log(p)$ approximates to $p$ for small numbers. Futhermore, assumptions about alternate guessing strategies have little effect on the predicted orderings of categorization utility as long as the strategy predicts that the receiver will do best when one attribute value is certain and will do worst when all attribute values are equally likely. In all our empirical applications to date, the most significant discrepancy between results of the information-theoretic and the feature-prediction versions are a few cases in which a tie in the goodness of two categories was broken by use of the other version.

## 4. APPLICATIONS TO CATEGORIZATION EXPERIMENTS

### 4.1 Murphy and Smith (1982)

Subjects in these experiments were taught nonsense names for a hierarchy of artificial categories with three levels of abstraction: subordinate, intermediate, and superordinate. In a later testing phase, subjects were shown a picture of a stimulus item, along with a category name, and were asked to indicate whether or not the stimulus was a member of the named category. The stimuli consisted of sixteen line drawings identified to subjects as examples of fictitious tools. The stimuli varied in their size (large or small) and in the shapes of their handles, shafts, and heads. This suggests a natural representation of the stimuli using four nominal attributes.

We evaluated the ability of Category Utility to predict which one of these levels of categorization is optimal. The optimal level is operationally defined to be the level at which people are quickest to verify that an object is a member of the category. According to our theory, the average Category Utility of the categories at a given level should be highest for this optimal level. We calculated both the average Category Utility($c$,F) of the individual categories and the average Category Utility(C,F) of the partitions induced by each individual category. For comparison, we also calculated the average cue validity, category validity, and the product of these two (G. Jones' (1983) "collocation" measure).

The intermediate level, the level that Murphy and Smith expected to be the basic level, indeed showed the fastest average name verification time, followed by the superordinate level. The subordinate level showed the longest reaction times. Cue validity, category validity, and Jones' collocation measure all failed to pick out this level as optimal - cue validity and the collocation measure selected the superordinate level, while category validity was constant across all levels. The average Category Utility(C,F) identified the basic and superordinate categories as equally good, with the subordinate level as worst. The average Category Utility(c,F) correctly identified the order of relative goodness as basic, superordinate, subordinate.

*4.2 Hoffmann und Ziessler (1983)*

Hoffmann and Ziessler (1983) replicated the basic level phenomena using three artificial category hierarchies. The hierarchies were differentiated by the degree to which exemplars of categories at different levels share common attribute values. Thus, a different level was expected to be basic in each of the hierarchies.

Subjects were assigned to learn one of the three hierarchies. They were taught to associate each item with category names at three levels of generality (e.g. exemplar, intermediate, superordinate). Following this, subjects were presented with a picture of one of the items, paired with a concept name. They were asked to verify, as quickly as possible, whether or not the picture was an example of the named category. In a second task, they were asked to recall the correct name at a given level of abstraction. Reaction times for both the verification and naming studies indicated that the basic level was at the superordinate level for one hierachy, at the intermediate level another, and at the exemplar level for the third hierarchy.

In these studies, cue validity and the collocation measure invariably identified the highest level as best. Category validity failed to distinguish between any of the levels. In summary, these measures were insensitive to the manipulation of attributes across the three hierarchies; each failed to predict the basic level in at least two out of three studies. The average Category Utility(c,F) correctly predicted the ordering of reaction times for the three levels in *each* of the three hierarchies, with the exception of giving equal ratings to the basic and intermediate levels in the first hierarchy. The average Category Utility(C,F) correctly predicted the ordering of reaction times in all three hierarchies.

## 5. DISCUSSION

The results from these experiments indicate that Category Utility is able to predict the psychologically preferred level of categorization in these verification and naming experiments. None of the alternative measures did nearly as well. An additional advantage to the measure is that it is context sensitive: Category Utility is computed as an expected *decrease* in uncertainty given some context population. Thus, this affords a way of measuring how the utility of a category or generalization can change depending on the context in which it is analyzed. This is particularly important from a psychological standpoint because of evidence indicating that the usefulness of categories and concepts is highly context dependent (Barsalou, 1982).

We have argued that the evidence indicates that psychologically preferred categories tend to be those that maximize potential information transfer. A number of clustering schemes have been advanced for finding cluster solutions that maximize information (e.g. Lance & Williams, 1966; Wallace & Boulton, 1968). More recent work that attempts to form prototype-based representations of categories while maximizing informational value has been done by Hanson (1985). We suggest that the results of such clustering programs may be particularly consistent with the types of categorizations made by humans, hence more explainable and valuable to users.

## References

Barsalou, L. W. (1982). Context-independent and context-dependent information in concepts. *Memory & Cognition, 10*, 82-93.

Berlin, B., Breedlove, D. E., & Raven, P. H. (1973). General principles of classification and nomenclature in folk biology. *American Anthropology, 75*, 214-242.

Daehler, M. W., Lonardo, R., & Bukatko, D. (1979). Matching and equivalence judgments in very young children. *Child Development, 50*, 170-179.

Hanson, S. J. (1985). Machine Learning, Clustering, and Polymorphy.. *Proceedings of Workshop on Uncertainty and Belief, Los Angeles.*

Hoffmann, J. & Ziessler, C. (1983). Objectidentifikation in kunstlichen Begriffshierarchien. *Zeitschrift fur Psychologie, 194*, 135-167.

Jolicoeur, P., Gluck, M., & Kosslyn, S. (1984). Pictures and names: Making the connection. *Cognitive Psychology, 16*, 243-275.

Jones, G. V. (1983). Identifying basic categories. *Psychological Bulletin, 92*, 174-177.

Lance, G. N. & Williams, W. T. (1966). Computer programs for hierarchical polythetic classification . *Computer Journal, 9*, 60-64.

Medin, D. (1983). Structural principles of categorization. In B. Shepp & T. Tighe (Eds.), *Interaction: Perception, Development and Cognition.* Hillsdale: Erlbaum.

Murphy, G. (1982). Cue validity and levels of categorization. *Psychological Bulletin, 91*, 174-77.

Murphy, G. L. & Smith, E. E. (1982). Basic level superiority in picture categorization. *Journal of Verbal Learning and Verbal Behavior, 21*, 1-20.

Rosch, E. & Mervis, C. (1975). Family resemblances: Studies in the internal structure of categories. *Cognitive Psychology, 7*, 573-603.

Rosch, E., Mervis, C., Gray, W., Johnson, D., & Boyes-Braem, P. (1976). Basic objects in natural categories. *Cognitive Psychology, 8*, 382-439.

Shannon, C. & Weaver, W. (1949). *The Mathematical Theory of Communication.* Chicago: University of Illinois Press.

Wallace, C. S. & Boulton, D. M. (1968). An information measure for classification. *Computer Journal, 11*, 185-194.